\documentclass[english]{DDCLSconf}
\usepackage[comma,numbers,square,sort&compress]{natbib}
\usepackage{epstopdf}
\usepackage{amsmath}
\usepackage{subfigure}
\usepackage{algorithm}
\usepackage{algorithmic}
\begin{document}

\title{Wasserstein Distance guided Adversarial Imitation Learning with Reward Shape Exploration}

\author{Ming Zhang\aref{sigs,}$^{\ast}$,
		Yawei Wang\aref{sigs,}$^{\ast}$,
		Xiaoteng Ma\aref{datu},
		Li Xia\aref{bsys,}$^{\dagger}$,
		Jun Yang\aref{datu,}$^{\dagger}$,
		Zhiheng Li\aref{sigs},
		Xiu Li\aref{sigs}
}

\affiliation[sigs]{The Shenzhen International Graduate School, Tsinghua University, Shenzhen 518055, P.~R.~China
        \email{zhangming\_0706@163.com},{wangyw0511@163.com},{zhhli@tsinghua.edu.cn},{li.xiu@sz.tsinghua.edu.cn}}
    
\affiliation[datu]{The Department of Automation, Tsinghua University, Beijing 100084, P.~R.~China
	\email{ma-xt17@mails.tsinghua.edu.cn},{yangjun603@tsinghua.edu.cn}}   
 
\affiliation[bsys]{The Business School, Sun Yat-Sen University, Guangzhou 510275, P.~R.~China
        \email{xiali5@sysu.edu.cn}}

\maketitle

\begin{abstract}
The generative adversarial imitation learning (GAIL) has provided an adversarial learning framework for imitating expert policy from demonstrations in high-dimensional continuous tasks. However, almost all GAIL and its extensions only design a kind of reward function of logarithmic form in the adversarial training strategy with the Jensen-Shannon (JS) divergence for all complex environments. The fixed logarithmic type of reward function may be difficult to solve all complex tasks, and the vanishing gradients problem caused by the JS divergence will harm the adversarial learning process. In this paper, we propose a new algorithm named Wasserstein Distance guided Adversarial Imitation Learning (WDAIL) for promoting the performance of imitation learning (IL). There are three improvements in our method: (a) introducing the Wasserstein distance to obtain more appropriate measure in adversarial training process, (b) using proximal policy optimization (PPO) in the reinforcement learning stage which is much simpler to implement and makes the algorithm more efficient, and (c) exploring different reward function shapes to suit different tasks for improving the performance. The experiment results show that the learning procedure remains remarkably stable, and achieves significant performance in the complex continuous control tasks of MuJoCo$^{1}$. 
\end{abstract}


\footnotetext{$\ast$ M. Zhang and Y. Wang contribute equally to this work.}
\footnotetext{$\dagger$ L. Xia and J. Yang are the corresponding authors. }
\footnotetext{${^1}$ Our code is available at https://github.com/mingzhangPHD/Adversarial-Imitation-Learning. }

\section{Introduction}

Reinforcement learning (RL) has been widely used in control problems of complex environments and it has made great breakthroughs in recent years \cite{mnih2015human,2_silver2016mastering}. The essential of RL is the Markov decision process (MDP) which aims at maximizing the cumulative rewards from the environment. The reward function is carefully hand-designed by the related expert before the agent learning the policy to solve the control task, and it is completely clear about the purpose of the control mission when the reward function is determined. However, when it comes to real-world applications, it will be quite difficult and time-consuming to design a reward function reasonable by hand. Imitation learning (IL) has the potential to overcome this challenge by learning how to perform tasks directly from expert state-action demonstrations. In general, imitation learning can be typically divided into two categories: \emph{behavioral cloning} (BC) and \emph{inverse reinforcement learning} (IRL). BC\cite{4_ross2011reduction} is the simplest method that casts the problem of imitation as supervised learning without environment interactions during training, and it is the first choice when enough demonstrations are available. 

However, BC often fails to imitate the expert behavior in real-world high-dimensional environments. In contrast, IRL \cite{8_ng2000algorithms} focuses on searching for a reward function that could best explain the demonstrated behavior. Yet the function search is ill-posed as the demonstration behavior may correspond to multiply reward functions, and the IRL methods inevitably suffer from the intensive computation which limits their efficiency in high-dimensional scenarios.

Generative adversarial imitation learning (GAIL) \cite{14_ho2016generative} has become a very popular model-free imitation learning framework for directly extracting policy from the expert demonstrations, which takes advantage of Generative Adversarial Network (GAN) \cite{15_goodfellow2014generative} and outperforms the BC and IRL in many complex applications. 

Nevertheless, there are still several limitations in GAIL which are much needed to be improved: (1) due to the problem of gradients vanishing incurred by the mathematical property of JS divergence in GAN, the reward signal from discriminator may be unstable \cite{16_li2019dialogue}, (2) the sample efficiency in terms of environment interactions and the learning speed during training are still not satisfactory \cite{6_sasaki2018sample,14_ho2016generative}, and (3) the reward function based on the discriminator is biased, which may lead to sub-optimal behaviors in some complex continuous environments \cite{17_kostrikov2018discriminator}.

In order to promote the existing limitations in GAIL, we propose a novel algorithm, called WDAIL: Wasserstein Distance guided Adversarial Imitation Learning. For the unstable problem in the adversarial learning process, we replace the origin JS divergence with the Wasserstein distance which has more reasonable mathematical property. Then, the PPO algorithm is used at the RL phase for making our method own better learning speed and sample efficiency and more easily to realize. Motivated by the bias problem in reward function discussed by Kostrikov \emph{et al.} \cite{17_kostrikov2018discriminator}, we explore different reward function shapes for reaching the high performance in different continuous control tasks. To the best of our knowledge, we are the first to take the different shapes of reward function into consideration in algorithms based on adversarial imitation learning. Regardless of the reward shapes, we prefer the reward function that can provide correct guidelines for policy improvement. In other words, we find that the reward shape does not have to be rigid in fixed logarithmic forms and their combinations. Several other sorts of reward function have been proved to achieve comparable or even better performance than GAIL's. The experimental results are introduced in detail in Section 4.

\section{Related Work}

Recently, some imitation learning algorithms combining with GAN have been proposed\cite{13_finn2016guided,14_ho2016generative}. In these methods, the Generative Adversarial Imitation Learning (GAIL) has achieved the  state-of-art performance in many typical applications \cite{23_li2017infogail,24_song2018multi}. 

However, it may be unstable and sample-inefficient when training GAIL.The training procedure of GAIL can be unstable because balancing the interplay between the discriminator and the generator is difficult. 
To solve this problem, some researchers proposed the alternative loss function with regularize items or added constraint condition on the discriminator \cite{21_arjovsky2017wasserstein,22_gulrajani2017improved,31_peng2018variational}. 
The previous work\cite{23_li2017infogail} has integrated WGAN into their InfoGAIL framework. However, they simply took WGAN as a trick without rigorous theoretical analysis and the task they concentrated on differs from ours as well. GAIL demands numerous interactions with the environment to reduce the variance of the Monte Carlo estimate of the state value, which extremely increases the sample complexity. Therefore, some recent  attempts \cite{6_sasaki2018sample,17_kostrikov2018discriminator} have been made to decrease the interactions.

Another attractive research field is exploring the reward function shape that generalizes better across all the RL tasks. Reward shapes used in previous methods are unstable or biased \cite{16_li2019dialogue,17_kostrikov2018discriminator}. The Adversarial Inverse Reinforcement Learning (AIRL) algorithm\cite{fu2017learning} uses the reward function: $r\left(s, a\right)=\log \left(D\left(s, a\right)-\log \left(1-D\left(s, a\right)\right)\right.$, which can assign both positive and negative rewards for each time step. But this kind of reward function may finish an episode earlier instead of trying to imitate the expert and lead to sub-optimal policies in environments with a survival bonus \cite{17_kostrikov2018discriminator}.
Kostrikov \emph{et al.} tried to solve the bias problem by redefining absorbing state reward. However, the results they demonstrated in their paper could not be completely reproduced by other researchers \cite{19_benardiclr}. Our work comes in such a situation to push forward studying the impact of reward function in adversarial learning.

\section{Proposed Method}
In this section, we detail the novel Wasserstein Distance guided Adversarial Imitation Learning(WDAIL) algorithm to solve the imitation learning problem in the high-dimensional continuous control tasks.

\subsection{Adversarial Imitation Learning with Wasserstein Distance}
The adversarial imitation learning algorithms are derived from the apprenticeship learning \cite{9_abbeel2004apprenticeship} which simply aims at matching occupancy measures with the distribution of expert demonstrations. These algorithms may be suitable for the small-scale problems, but has limited applicability. 

In these approaches, people often carefully design the reward function space $\mathcal{R}$ for deriving the model more effective and robust. The early methods mainly conducted the linear hand-engineered reward function \cite{9_abbeel2004apprenticeship,11_ramachandran2007bayesian},and deep models are applied to learn the non-linear function in recent approaches\cite{14_ho2016generative,16_li2019dialogue,17_kostrikov2018discriminator}. 

The apprenticeship learning approach through IRL aims at imitating a policy $\pi$ closing to the unknown expert policy $\pi_{E}$ while corresponding to occupancy measure condition $\rho_{\pi}:=\rho_{\pi_{E}}$. The optimization problem formulation of such an algorithm is defined as follows:
\begin{equation}
\arg \min _{\pi \in \Pi}\left\{-H(\pi)+\sup _{c \in \mathcal{R}} E_{\pi_{E}}[c(s, a)]-E[c(s, a)]\right\}.
\end{equation}
The Generative Adversarial Imitation Learning (GAIL) algorithm has been proposed for adapting the complex environment, the optimization problem in equation (1) is relaxed into the following form:
\begin{equation}
\min _{\pi \in \Pi}\left\{ \psi\left(\rho_{\pi}-\rho_{\pi_{E}}\right)-H(\pi)\right\},
\end{equation}
where the $\psi(\rho_{\pi}-\rho_{\pi_{E}})$ denotes the distribution discrepancy of the occupancy measures between the distribution of policy and expert trajectories. 
In GAIL, the choice of $\psi$ is inspired by the Generative adversarial networks (GAN) \cite{15_goodfellow2014generative}, which is the following fact:
\begin{equation}
\begin{split}
\psi(\rho_{\pi}-\rho_{\pi_{E}})&= \max _{D \in(0,1)^{S \times A}} \left\{E_{\pi}[\log (D(s, a))]\right.\\&\left.+E_{\pi_{E}}[\log (1-D(s, a))]\right\}.
\end{split}
\end{equation}

The equation presents the maximum term of the GAN which is an optimization problem of binary classification, the purpose of this form is to distinguish that the state-action pairs come from policy $\pi$ or expert policy $\pi_{E}$. Based on the theoretical analysis of the GAN, such an objective function is the JS divergence between two different distributions:
\begin{equation}
\begin{split}
D_{J S}\left(\rho_{\pi}, \rho_{\pi_{E}}\right) &=  D_{KL}(\rho_{\pi}\parallel(\rho_{\pi}+\rho_{E})/2)\\
&+D_{KL}(\rho_{\pi_{E}}\parallel(\rho_{\pi}+\rho_{E})/2).
\end{split}
\end{equation}
With the policy causal entropy item $H(\pi)$, the final GAIL algorithm is developed as follows:
\begin{equation}
\min _{\pi \in \Pi} \left\{D_{J S}\left(\rho_{\pi}, \rho_{\pi_{E}}\right)-H(\pi)\right\}.
\end{equation}
From the formulation, the KL divergence is regarded as the occupancy measure and the JS divergence is responsible for measuring the true metric between the distributions of the policy and expert demonstrations.

Based on the discussion above, we consider that the distribution measure function $\psi$ can be replaced by the $L_1$-Wasserstein distance. $L_1$-Wasserstein distance is a reasonable way to define a proper distance metric between two distributions, and it is continuous and differentiable almost everywhere. For any two probability measures $P$ and $Q$ on $M$, the Wasserstein distance of order $p$ is defined as follows:
\begin{equation}
W_{p}(P, Q)=\left(\inf _{\pi \in \Pi(P, Q)} \int_{M} \rho(x, y)^{p} d \pi(x, y)\right)^{\frac{1}{p}},
\end{equation}
where $\rho(x,y)$ is a distance function; $x$ and $y$ are the samples from set $S$; $\prod(P,Q)$ is the set of all probability measures on $S\times S$ with marginals $P$ and $Q$. 
The $L_1$-Wasserstein distance is more flexible and easier to bound and it has strong implications in functional analysis.
The Kantorovich-Rubinstein duality tells us that, the $L_1$-Wasserstein distance  (Earth-Mover distance) could be expressed as follows: 
\begin{equation}
W_{1}(P, Q)=\sup _{\|f\| \leq 1}\left\{ E_{x \sim P}[f(x)]-E_{x \sim Q}[f(x)]\right\},
\end{equation}
where the Lipschitz semi-norm is $\| f\|=\mathrm{sup}|f(x)-f(y)|/\rho(x,y)$. 

In this paper, we take $L_1$-Wassertain distance into the adversarial imitation learning procedure. According to equation (7), the objective of Wasserstein Distance guild Adversarial Imitation Learning (WDAIL) can be formulated as follows:
\begin{equation}
\min _{\pi \in \Pi}\left\{-H(\pi)+W_{1}^{d}\left(\rho_{\pi}, \rho_{E}\right)\right\},
\end{equation}
where $W^d_1$ is the $L_1$-Wasserstein distance between the occupancy measure $\rho_{\pi}$ and $\rho_{E}$ drawing from the policy and expert, and $d$ is a valid distance metric defined in the state-action space $\mathcal{S} \times \mathcal{A}$.
In this formulation, it will be appropriate for large-scale environments through the gradient optimization approach.
The Wasserstein distance $W^d_1$ in equation (8) can be interpreted in the following form:
\begin{equation}
W_{1}^{d}\left(\rho_{\pi}-\rho_{\pi_{E}}\right)=\max _{\|d\|_{L} \leq 1} \left\{E_{\pi_{E}}[d(s, a)]-E_{\pi}[d(s, a)]\right\},
\end{equation} 
where the measure function $d(s, a)$ should be restrained by the 1-Lipschitz condition. 
In practice, the weight clipping \cite{21_arjovsky2017wasserstein} and gradient penalty \cite{22_gulrajani2017improved} are the effective methods for enforcing the $L_1$-Wasserstein distance satisfying the 1-Lipschitz constraint.

\subsection{Policy Optimization} 
We take advantage of the proximal policy optimization (PPO) algorithm \cite{35_RN2231} in our WDAIL instead of the trust region policy optimization (TRPO) \cite{36_RN2239} used in GAIL, which is much easier to implement, more general, and has better sample complexity.
The loss objective of the PPO algorithm is the surrogate item with a little change to the typical policy gradient (GP) algorithms. The implementation of the objective function is to construct the loss $L_t^{CLIP}$ or $L^{KLPEN}$ to substitute $L^{PG}$, and then perform multiple steps of stochastic gradient ascent on this objective. In this work, we make use of the clipped version of objective $L^{CLIP}$ which performs best in the comparison results \cite{35_RN2231}. The objective is expressed as follows:
\begin{small}
\begin{equation}
L^{C L I P}(\theta)=\hat{E}\left[\min \left(r_{t}(\theta) \hat{A}_{t}, {clip}\left(r_{t}(\theta), 1-\epsilon, 1+\epsilon\right) \hat{A}_{t}\right)\right],
\end{equation}
\end{small}where $r_t(\theta)$ denotes the probability ratio $r_t(\theta)=\frac{\pi_{\theta}(a_t|s_t)}{\pi_{\theta_{old}}(a_t|s_t)}$;
$\epsilon$ is a hyperparameter and $\hat{A}_{t}$ denotes the advantage estimate. With the clipping constraint, the policy will promote or decline into a certain boundary, this can make the optimization of policy performance more stable and reliable.

\begin{figure}
	
	\centering
	\subfigure{
		\begin{minipage}{0.6\textwidth}
			\centering
			\includegraphics[width=\textwidth]{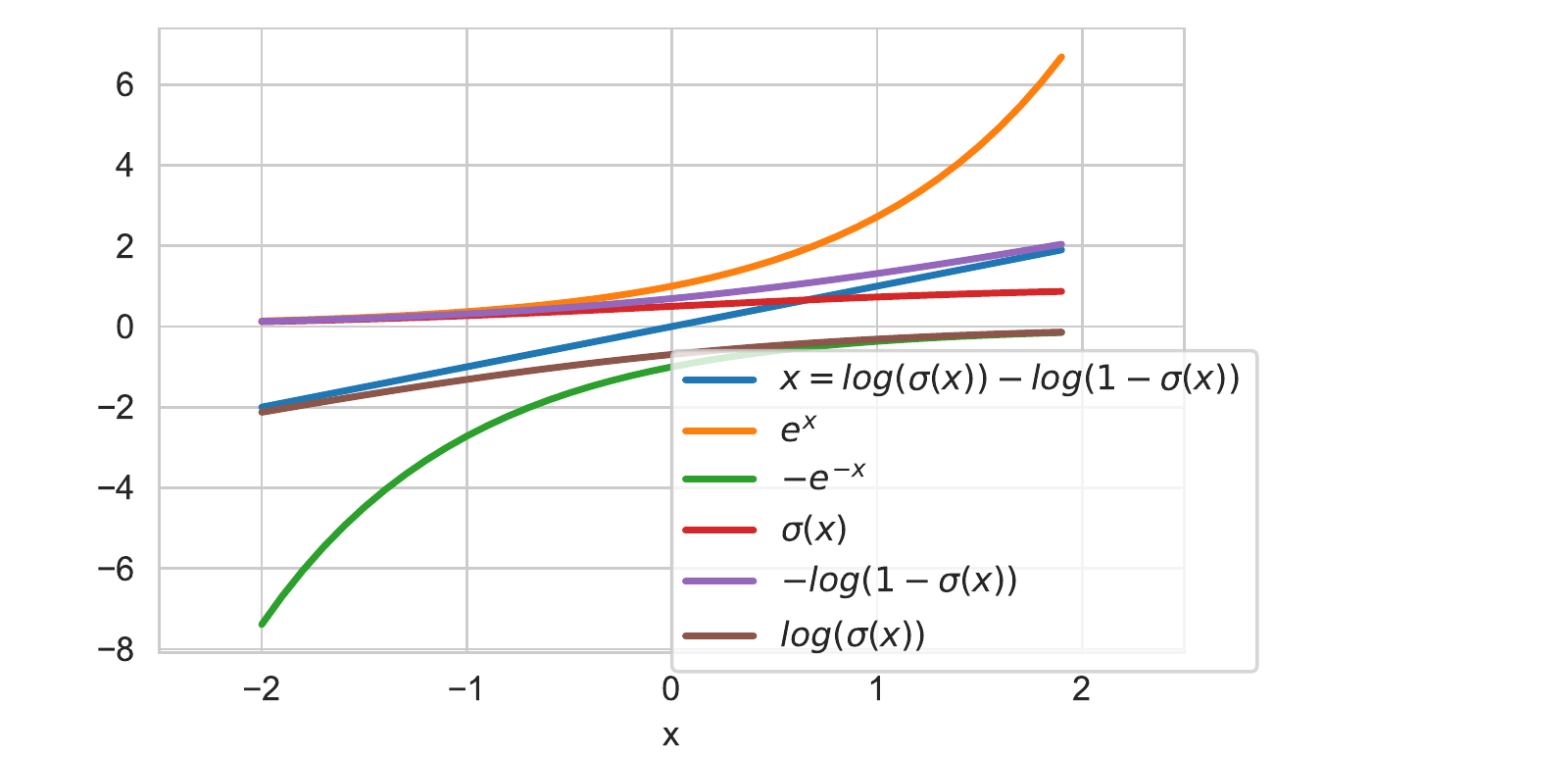}
		\end{minipage}
	}
	\caption{Different shapes of reward function.}
	\label{Fig_reward_functionss}
\end{figure}

\begin{figure*}[h]
	
	\centering
	\subfigure[Hopper]{
		\begin{minipage}{0.262\textwidth}
			\centering
			\includegraphics[width=\textwidth]{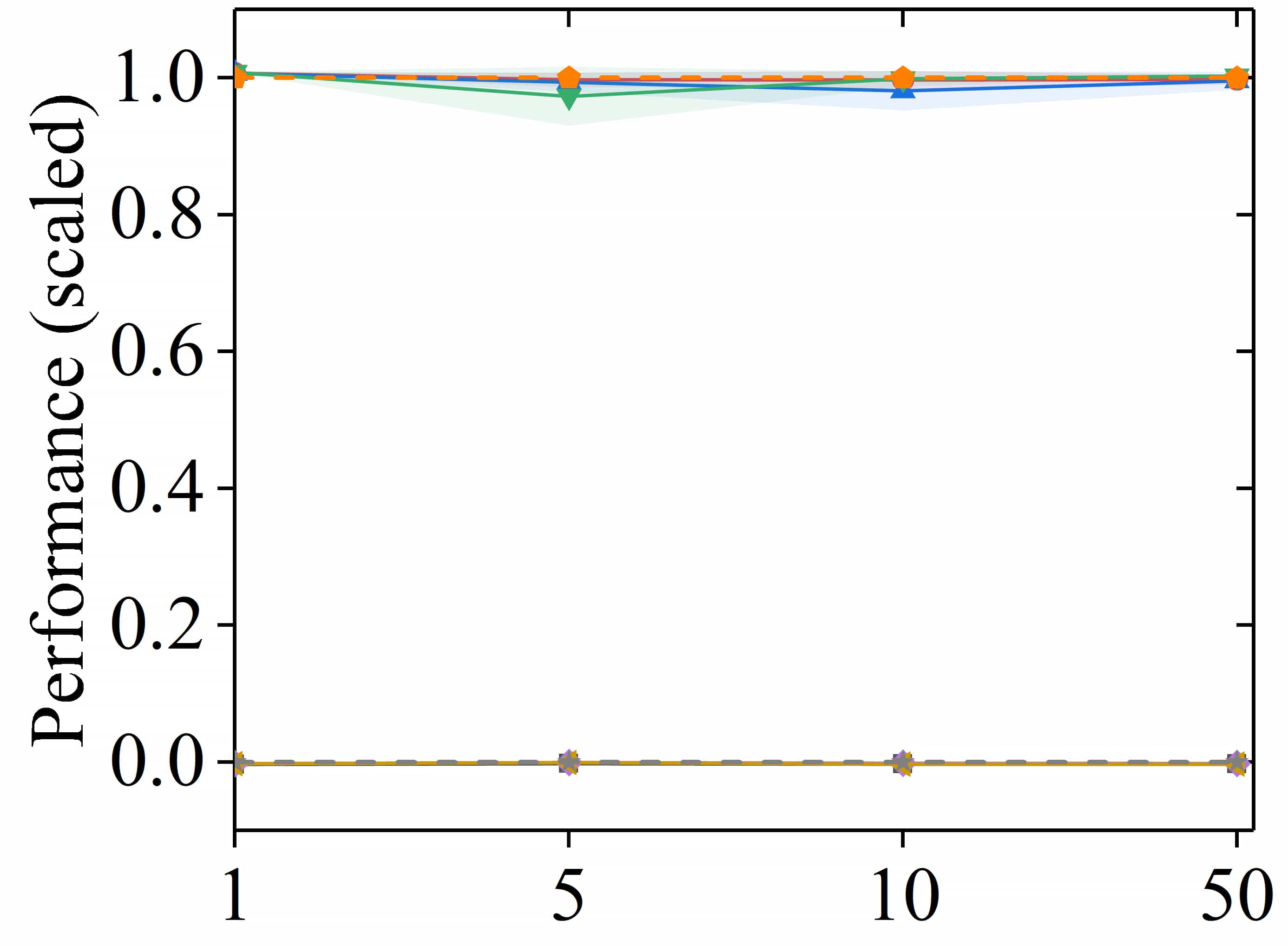}
			%			\caption{(a)}
		\end{minipage}
	}%
	\subfigure[Walker2d]{
		\begin{minipage}{0.24\textwidth}
			\centering
			\includegraphics[width=\textwidth]{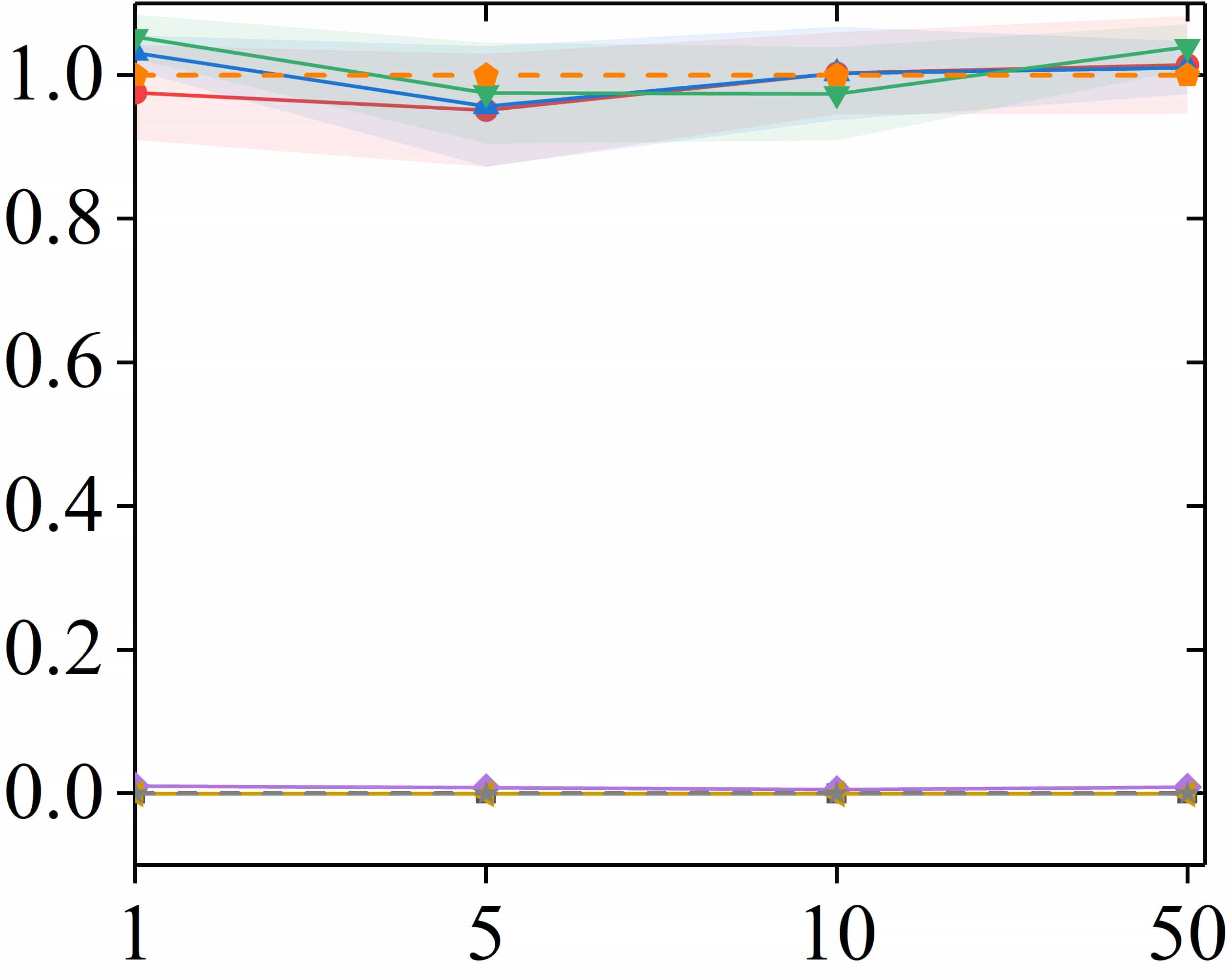}
			%			\caption{(a)}
		\end{minipage}
	}%
	\subfigure[HalfCheetah]{
		\begin{minipage}{0.24\textwidth}
			\centering
			\includegraphics[width=\textwidth]{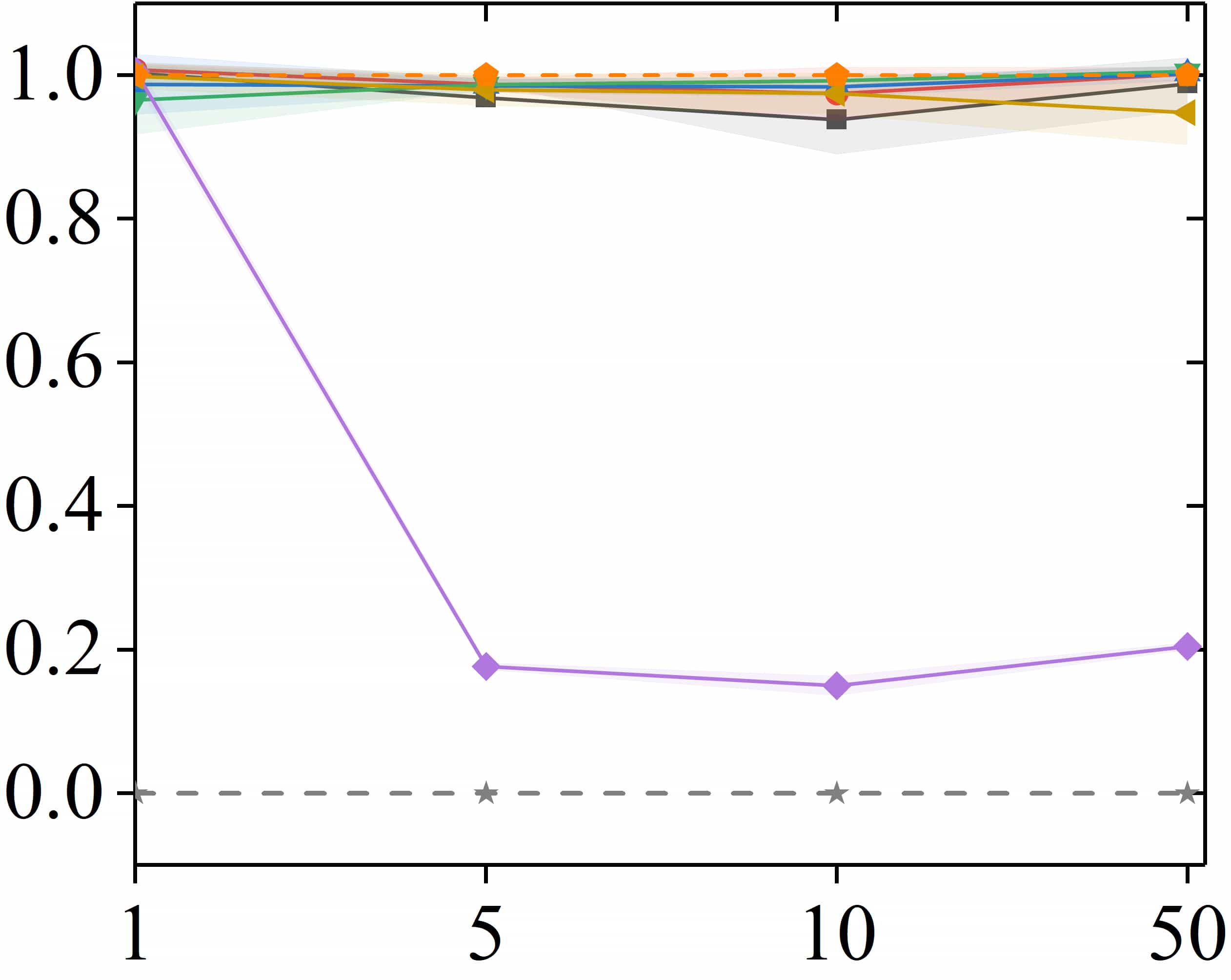}
			%			\caption{(a)}
		\end{minipage}
	}%
	\subfigure{
		\begin{minipage}{0.24\textwidth}
			\centering
			\includegraphics[width=\textwidth]{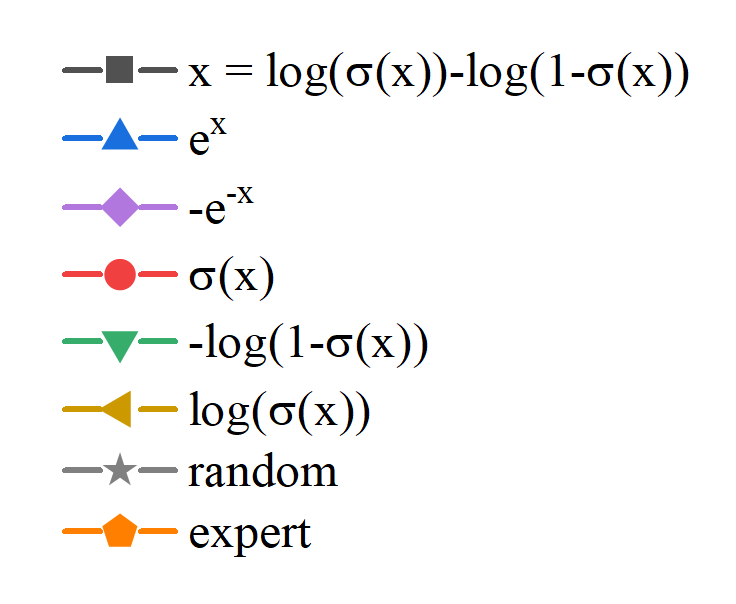}
			%			\caption{(a)}
		\end{minipage}
	}%
	
	\caption{Performances of learned policies using WDAIL with different reward shapes. The x-axis represents the sizes of expert trajectories.}
	\label{Fig_exp1}
\end{figure*}

\begin{figure*}
	\centering
	\subfigure[Hopper]{
		\begin{minipage}{0.26\textwidth}
			\centering
			\includegraphics[width=\textwidth]{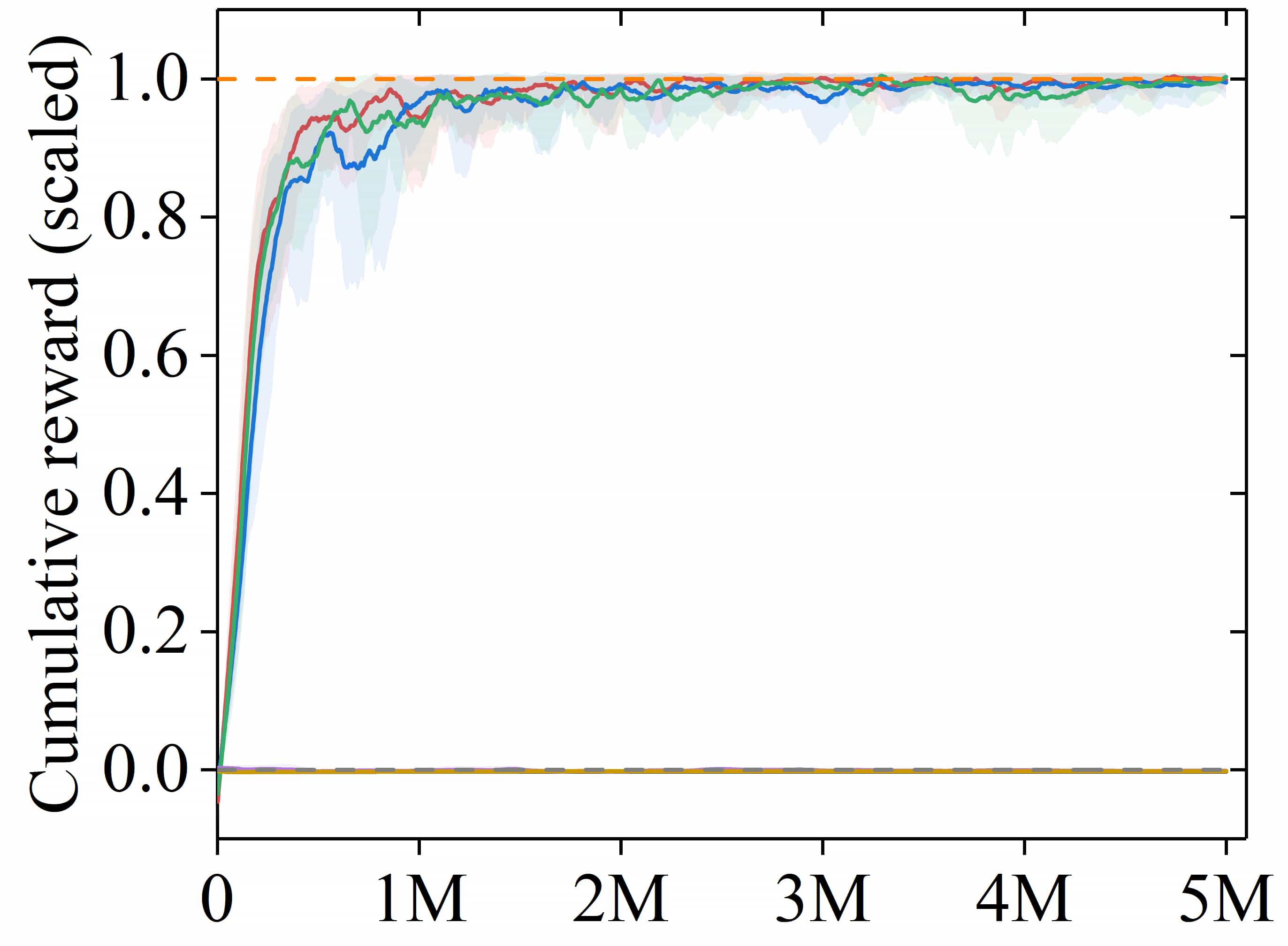}
			%			\caption{(a)}
		\end{minipage}
	}%
	\subfigure[Walker2d]{
		\begin{minipage}{0.241\textwidth}
			\centering
			\includegraphics[width=\textwidth]{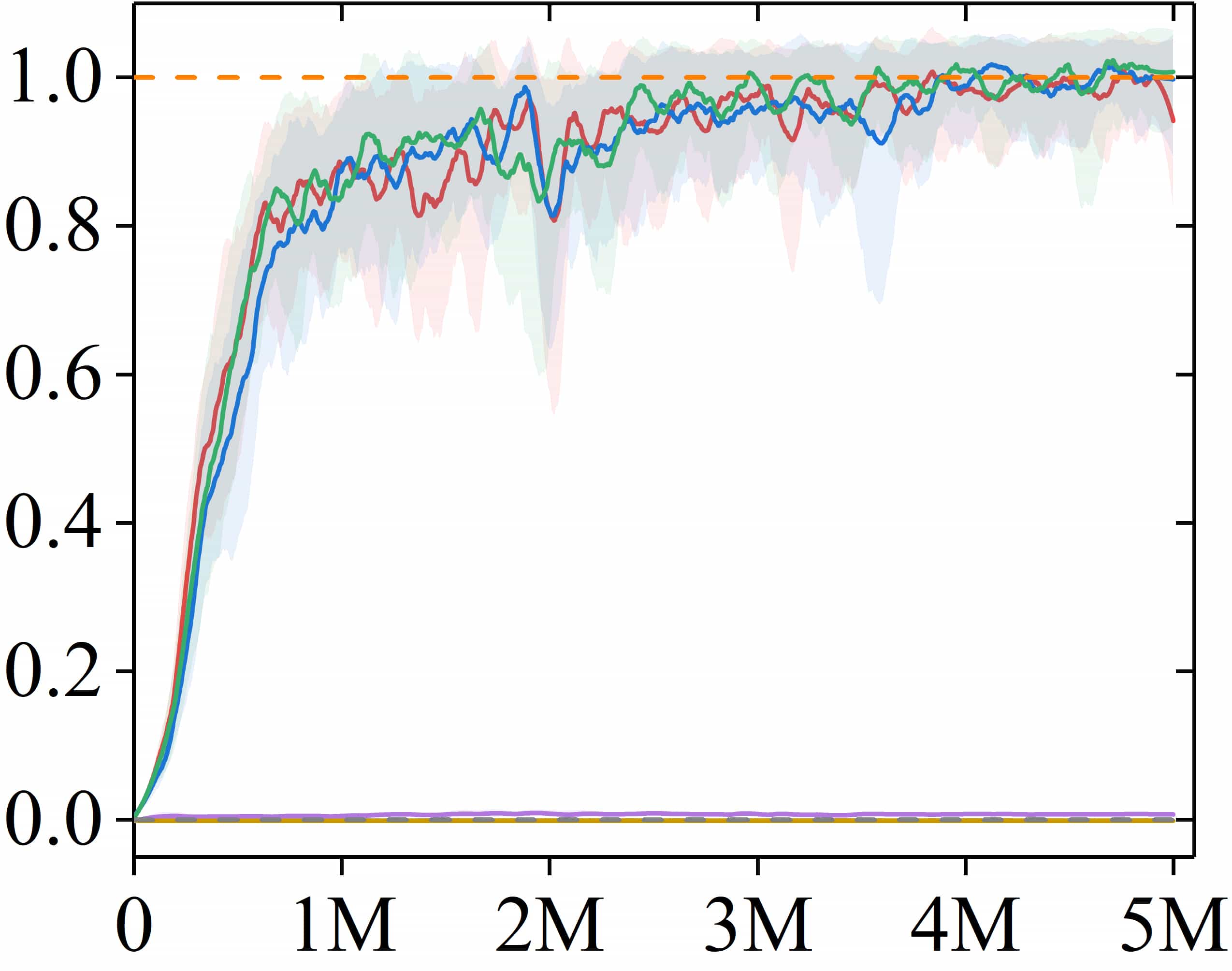}
			%			\caption{(a)}
		\end{minipage}
	}%
	\subfigure[HalfCheetah]{
		\begin{minipage}{0.246\textwidth}
			\centering
			\includegraphics[width=\textwidth]{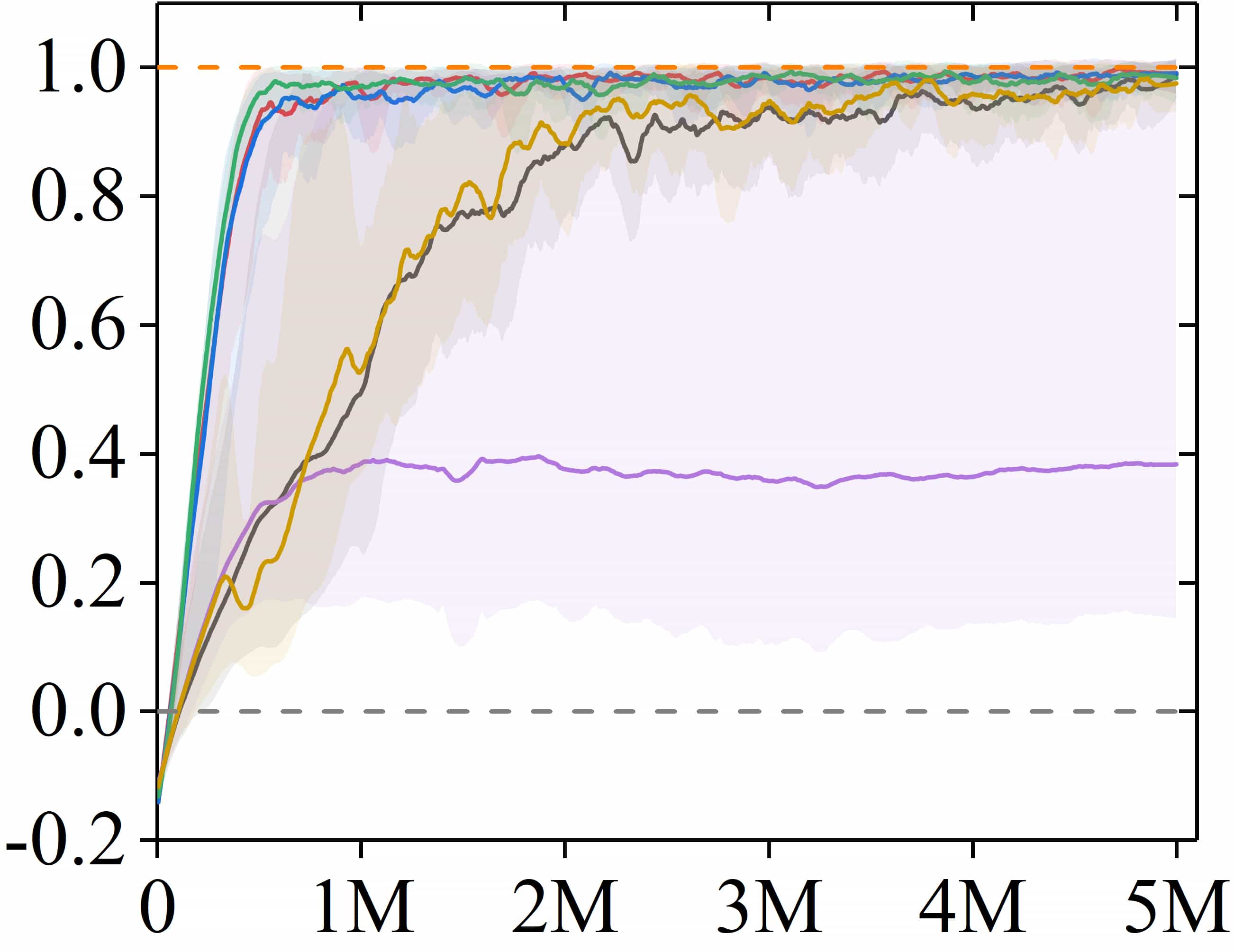}
			%			\caption{(a)}
		\end{minipage}
	}
	\subfigure{
		\begin{minipage}{0.212\textwidth}
			\centering
			\includegraphics[width=\textwidth]{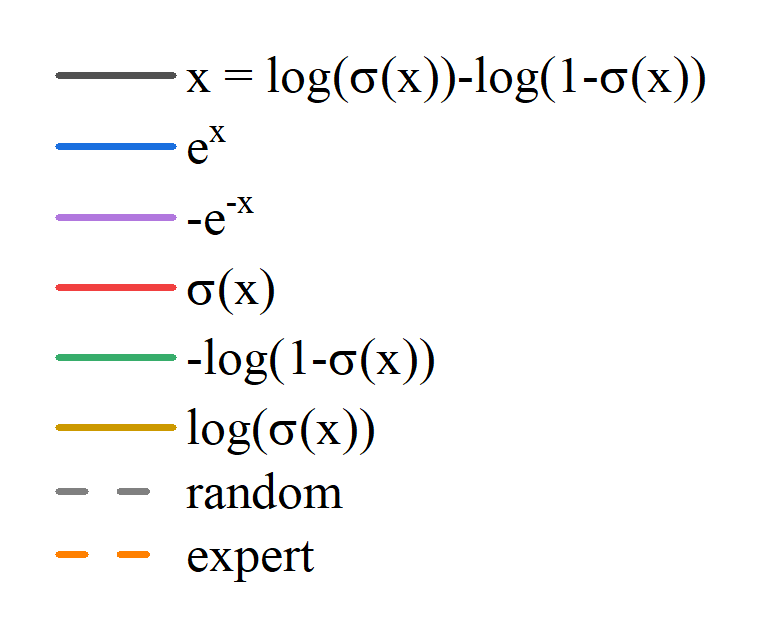}
			%			\caption{(a)}
		\end{minipage}
	}%
	\caption{Training curves of WDAIL with different reward shapes for different sizes of expert trajectories. The x-axis represents the times of interaction with the environment.}
	\label{Fig_further_exp1}
\end{figure*}

\subsection{Reward Shaping}
\label{reward shaping}
In the GAIL framework, the bias problem in the reward function has been paid attention to. The positive reward provides the survival bonus in each step in MDP, which motivates the agent to live longer. On the contrary, the negative form of reward will be the penalty in each step for a certain task. Intuitively, we can combine the positive and the negative reward in some formulation to reach a certain balance, what has been done in AIRL \cite{fu2017learning}. However, in the reproducibility of DAC \cite{19_benardiclr}, the unbiased reward function shape maybe not applicable to all MuJoCo environments. Through these ideas, we consider the shape of reward may be another critical component for learning the policy in the adversarial imitation learning algorithm.   
Therefore, we designed different shapes of the reward function, which are presented in Fig.~\ref{Fig_reward_functionss}. 
The input $x$ of these equations denotes the output of the discriminator $D(s,a)$, the $\sigma(x)=\frac{1}{1+e^{-x}}$ is the sigmoid function.
The direct output of discriminator $x$ is the linear reward shape, which is equal to the unbiased reward formulation $log(\sigma(x))-log(1-\sigma(x))$ proposed in \cite{17_kostrikov2018discriminator}.
Then the $x$ is clipped by the sigmoid function and a positive reward function $\sigma(x)$ is obtained.
Based on these two reward shapes, we combine them with exponential and logarithmic functions to create other reward functions, including the positive reward shapes $e^{x}$ and $-log(1-\sigma(x))$ and the negative reward shapes $-e^{-x}$ and $log(\sigma(x))$.

\subsection{Algorithm and Training Strategy}
In this work, we try to utilize the Wasserstein distance to substitute the JS divergence in GAIL, which is continuous and differentiable almost everywhere. The Wasserstein distance between the policy distribution $ P_{\tau_{\pi}} $ and expert distribution $ P_{\tau_E} $ can be defined in the following equation with the discriminator network $D$:
\begin{equation}
L_{wd}=\sum_{(s,a)^{E} \in \tau_E}D((s,a)^{E})-\sum_{(s,a)^{\pi} \in \tau_{\pi}}D((s,a)^{\pi}),
\end{equation}
where $(s,a)^{\pi}$ and $(s,a)^{E}$ are the state-action pairs drawing from policy transitions buffer $\mathcal{B}_{\tau_{\pi}}$ and expert transitions dataset $\mathcal{B}_{\tau_E}$, respectively. For satisfying the Lipschitz constraint condition of Wasserstein distance, Gulrajani \emph{et al.} \cite{22_gulrajani2017improved} proposed a more rational approach to add the constraint on the gradient of the discriminator network. The gradient penalty $L_{gp}((\hat{s},\hat{a}))$ can be expressed as follows:
\begin{equation}
L_{g p}=\left(\left\|\nabla_{(\hat{s}, \hat{a}) \in P_{(s, \hat{a})}} D((\hat{s}, \hat{a}))\right\|_{2}-1\right)^{2},
\end{equation}
where $(\hat{s},\hat{a})$ are the random samples from $P_{(\hat{s},\hat{a})}$, which are sampled uniformly along straight lines between pairs of points drawing from the policy and expert trajectories distribution. 

During the imitation learning process, the policy is promoted guided by Wasserstain distance through adversarial training strategy. Firstly, we train the discriminator by maximizing the Wasserstein distance between policy and expert trajectories and then minimize the Wasserstein distance through promoting the policy with the parameter fixed discriminator. The adversarial imitation learning strategy can be defined as a minimax game:
\begin{equation}
\underset{\theta}{min} \  \underset{w}{max} \  L_{wdail}(\theta,w)=  L_{wd}-\lambda L_{gp},
\end{equation}
where $\lambda$ is the penalty coefficient, $\theta$ and $w$ represent the parameters of policy and discriminator network, respectively. After the adversarial iteration, the optimal policy owning the ability to generate the most similar trajectories with expert trajectories will be ready, when the Wasserstein distance closes to zero. 
The algorithm of our method is summarized in Algorithm \ref{alg_1}.

\begin{algorithm}
	\caption{Wasserstein Distance guided Adversarial Imitation Learning}
	\label{alg_1}
	\textbf{Parameter}: $w$ and $\theta$
	\begin{algorithmic}[1]
		%		\REQUIRE Expert transitions buffer $\mathcal{B}_E$
		
		\FOR{$episode=1,2,...$}
		
		\FOR{$t=1,2,...,T$}
		
		\STATE Select action $a_t\sim \pi_{\theta_{old}}$ according to the old policy and current state $s_t$
		\STATE Act $a_t$ in environment and obtain next state $s_{t+1}$ 
		\STATE Store transition $(s_t, a_t)$ in transitions buffer $\mathcal{B}_{\pi}$
		\ENDFOR
		\STATE Update the discriminator parameters
		\FOR{$j=1,...,J$}
		
		\STATE Sample $\{(s_{(i)},a_{(i)})^{\pi}\}_{i=1}^m$ and $\{(s_{(i)},a_{(i)})^{E}\}_{i=1}^m$ from policy transitions buffer $\mathcal{B}_{\pi}$ and expert transitions $\mathcal{B}_E$
		\STATE Generate a random number $\epsilon \sim U[0,1]$
		\STATE $(\hat{s}, \hat{a})\gets \epsilon (s,a)^{\pi} +(1-\epsilon)(s,a)^{E}$
		\STATE $w\gets w-\alpha_d \nabla_w L_{wdail}$
		\ENDFOR
		\STATE Generate reward $r_t$ based on discriminator $D_{w}((s_t,a_t))$ and reward function $f_{rew}$:
		\FOR{$t=1,...,T$}
		\STATE $r_t=f_{rew}(D_w((s_t,a_t)))$
		\STATE Store $r_t$ in policy transitions buffer $\mathcal{B}_{\pi}$ ($s_t,a_t,r_t$)
		
		\ENDFOR	
		\STATE Compute advantage $\hat{A}$  and store in buffer $\mathcal{B}_{\pi}$ $(s,a,r,\hat{A})$
		\FOR{$k=1,...,K$}
		\STATE Sample mini-batch $\{ (s_i,a_i,r_i,A_i)^{\pi} \}^{m}_{i=1}$ from policy transitions buffer $\mathcal{B}_{\pi}$
		\STATE $\theta \gets \theta - \alpha_p \nabla_{\theta} L_t^{CLIP}$
		\STATE $\theta_{old}$ $\gets$ $\theta$
		\ENDFOR
		
		\ENDFOR
		
	\end{algorithmic}
\end{algorithm}

\begin{figure*}[h]
	
	\centering
	\subfigure[Hopper]{
		\begin{minipage}{0.259\textwidth}
			\centering
			\includegraphics[width=\textwidth]{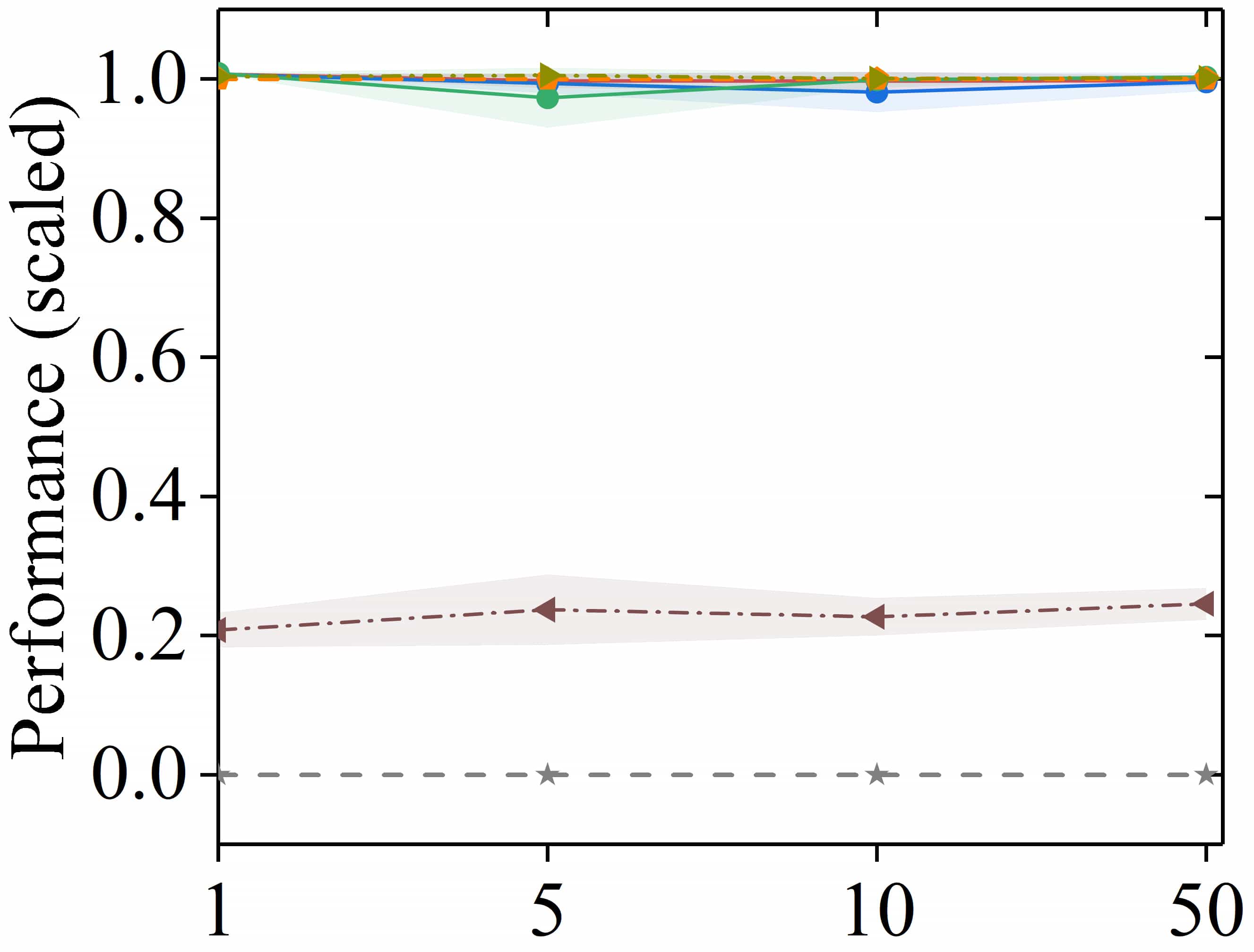}
			%			\caption{(a)}
		\end{minipage}
	}%
	\subfigure[Walker2d]{
		\begin{minipage}{0.24\textwidth}
			\centering
			\includegraphics[width=\textwidth]{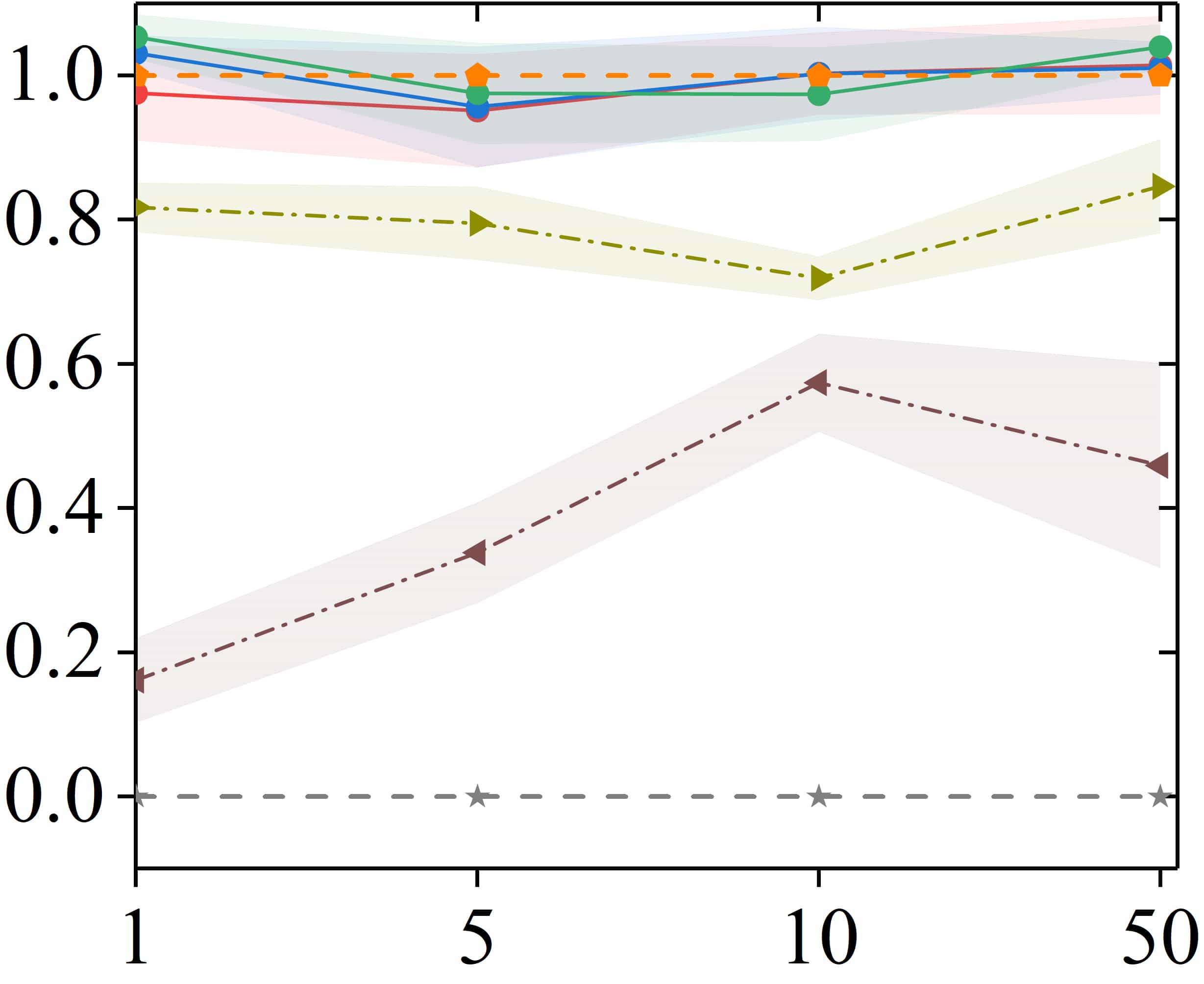}
			%			\caption{(a)}
		\end{minipage}
	}%
	\subfigure[HalfCheetah]{
		\begin{minipage}{0.243\textwidth}
			\centering
			\includegraphics[width=\textwidth]{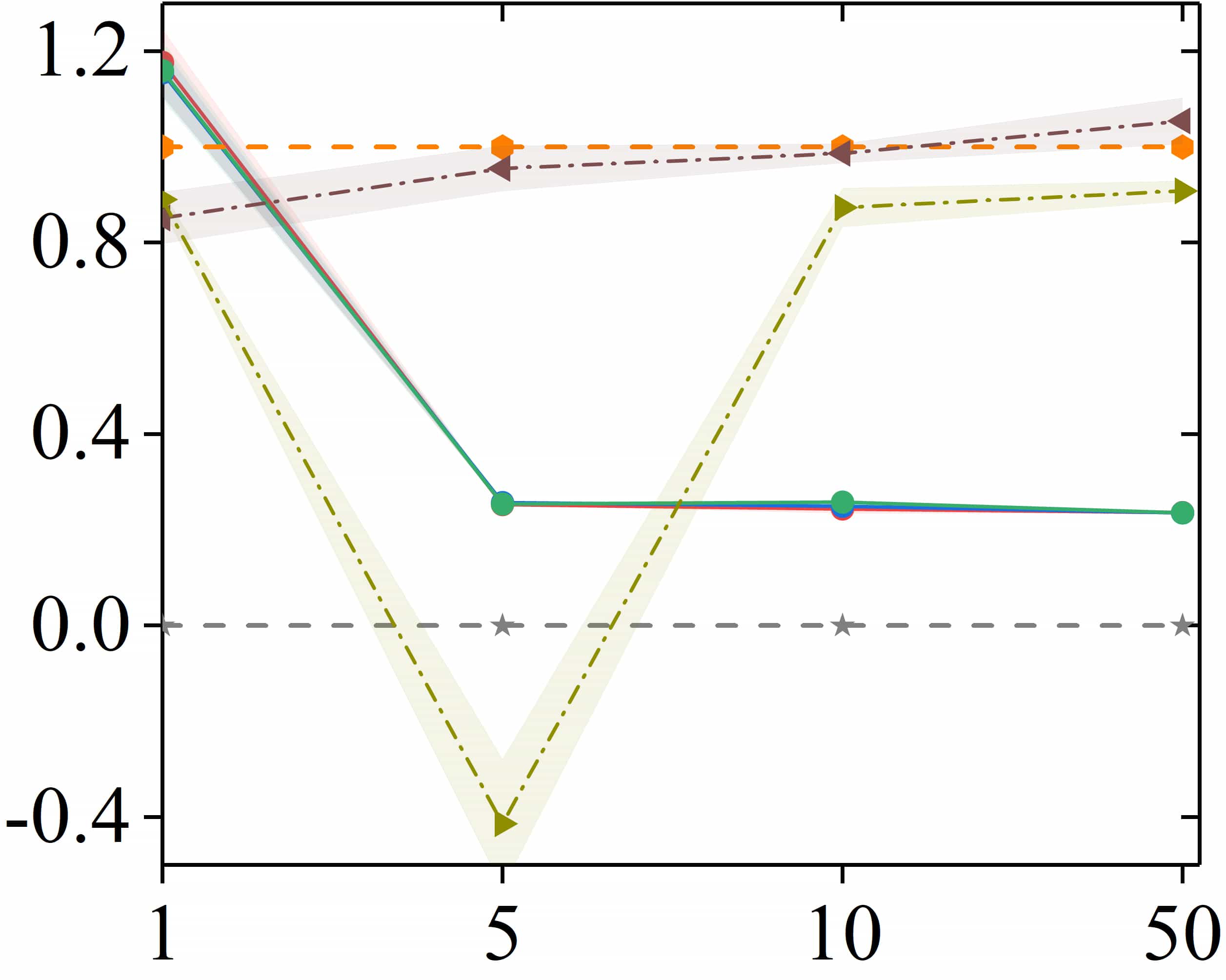}
			%			\caption{(a)}
		\end{minipage}
	}%
	\subfigure{
		\begin{minipage}{0.24 \textwidth}
			\centering
			\includegraphics[width=\textwidth]{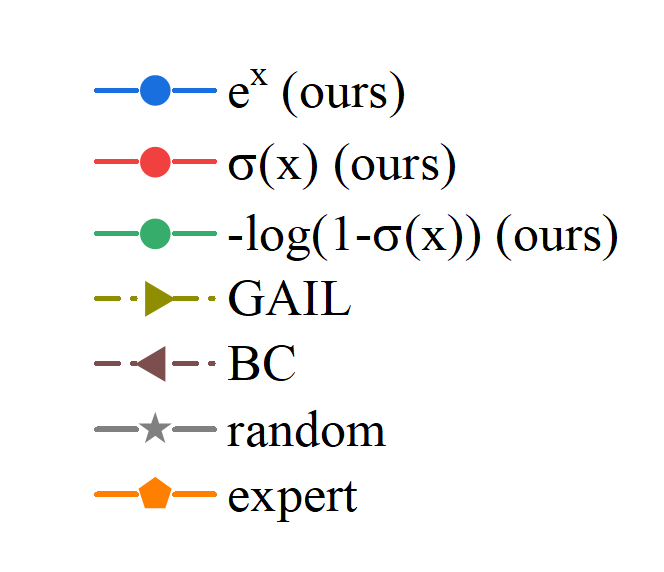}
			%			\caption{(a)}
		\end{minipage}
	}%
	\caption{Performances of learned policies using different algorithms. The x-axis represents the sizes of expert trajectories.}
	\label{Fig_exp2}
\end{figure*}

\begin{figure*}
	\centering
	\subfigure[Hopper]{
		\begin{minipage}{0.262\textwidth}
			\centering
			\includegraphics[width=\textwidth]{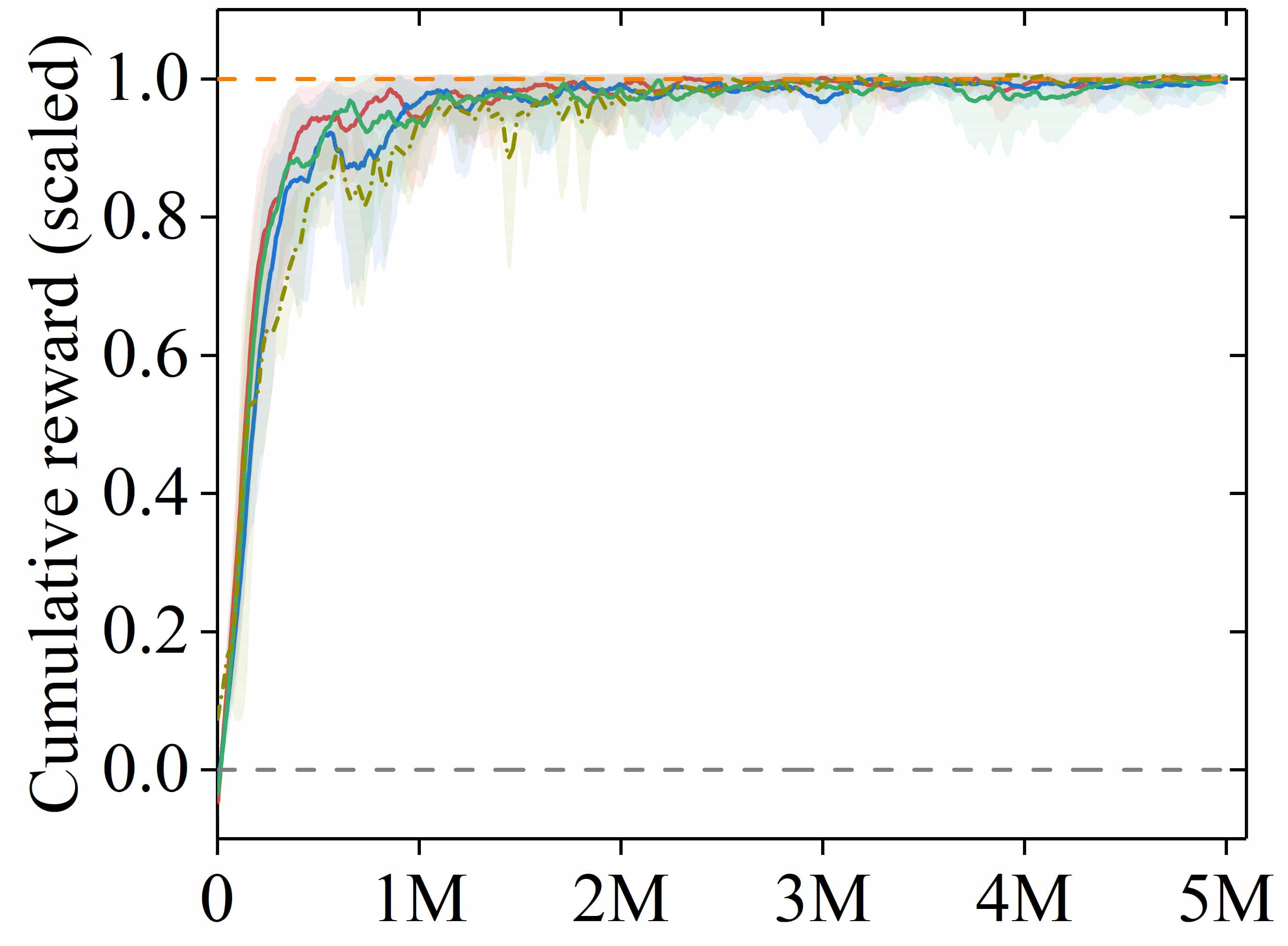}
			%			\caption{(a)}
		\end{minipage}
	}%
	\subfigure[Walker2d]{
		\begin{minipage}{0.245\textwidth}
			\centering
			\includegraphics[width=\textwidth]{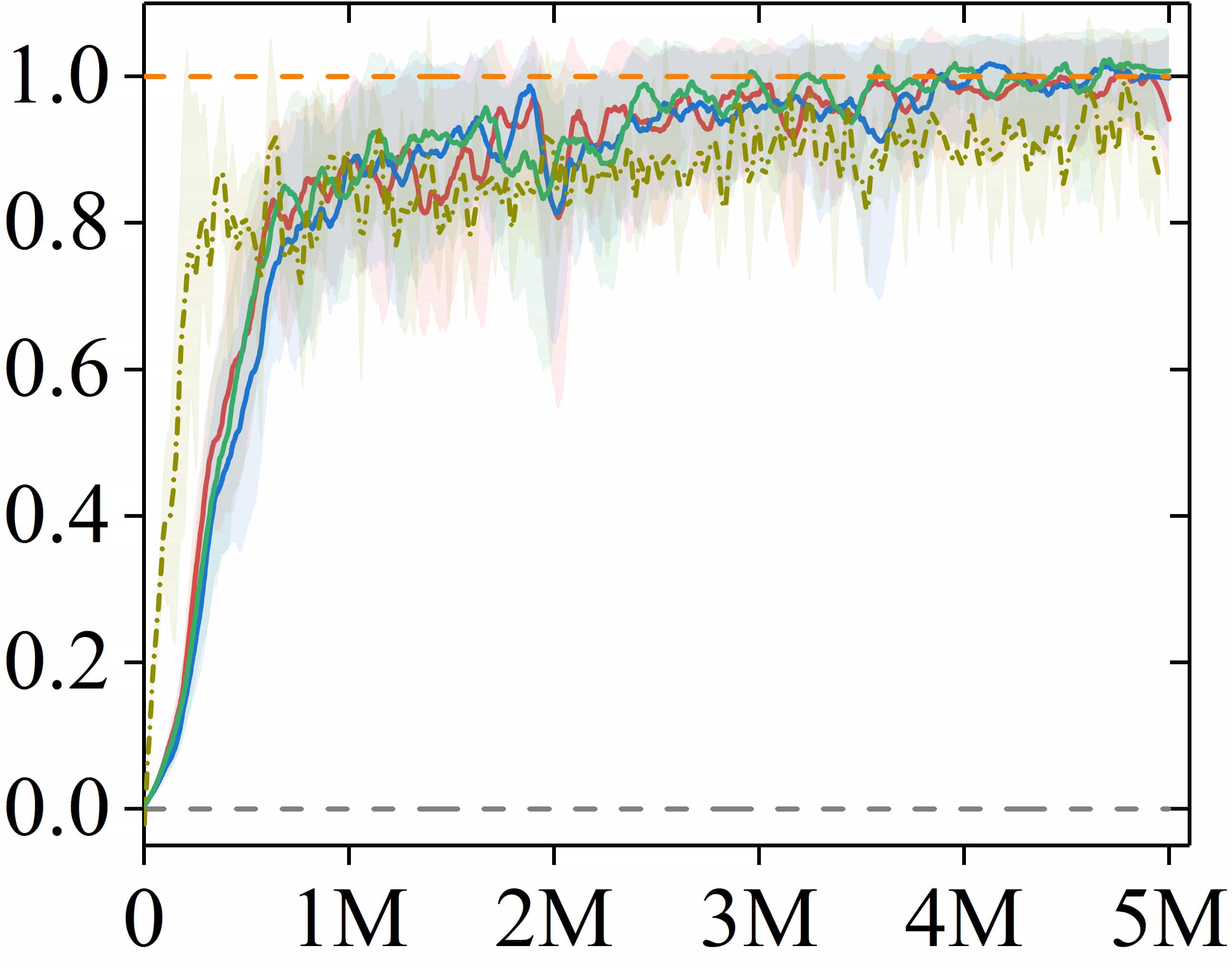}
			%			\caption{(a)}
		\end{minipage}
	}%
	\subfigure[HalfCheetah]{
		\begin{minipage}{0.256\textwidth}
			\centering
			\includegraphics[width=\textwidth]{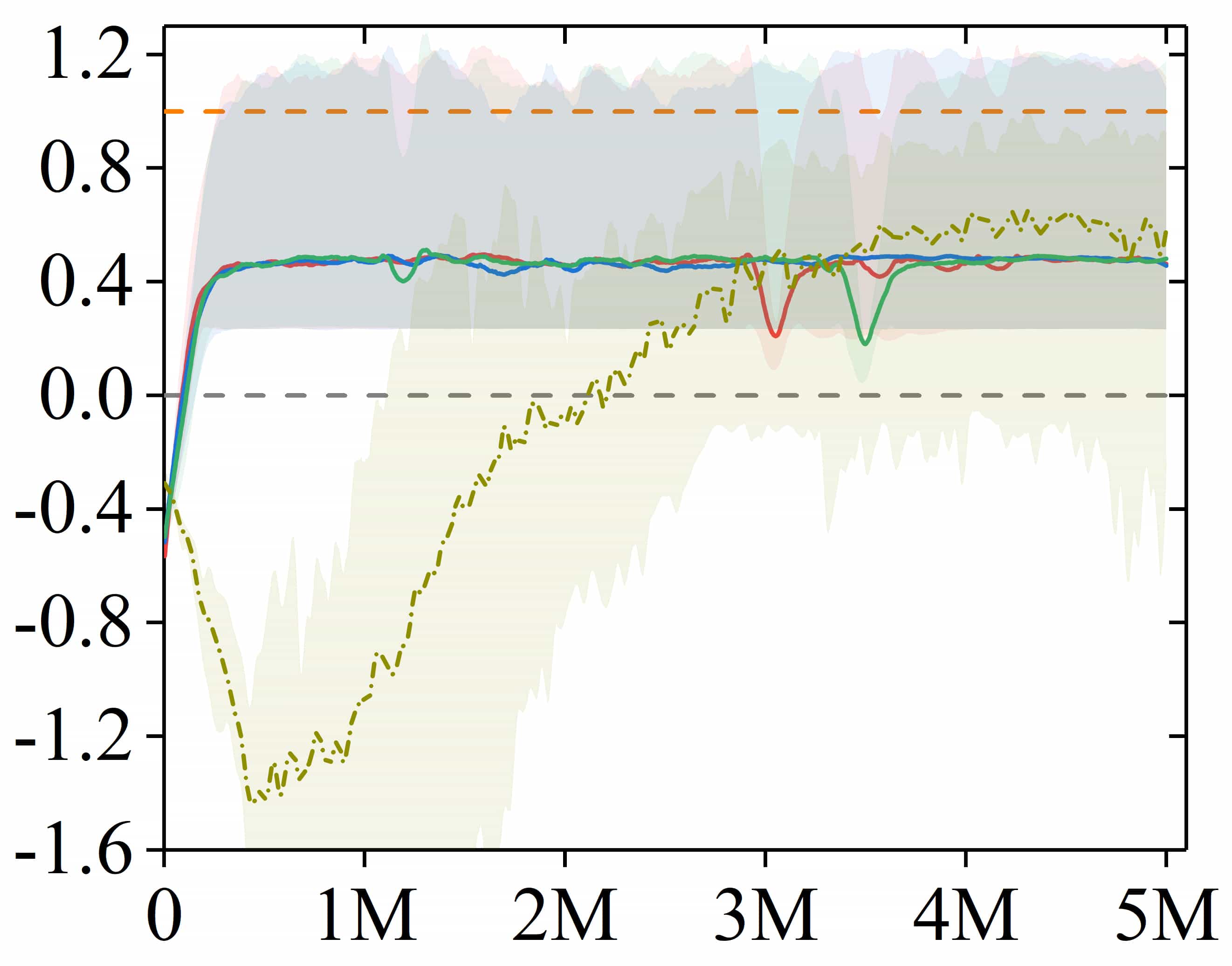}
			%			\caption{(a)}
		\end{minipage}
	}%
	\subfigure{
		\begin{minipage}{0.24\textwidth}
			\centering
			\includegraphics[width=\textwidth]{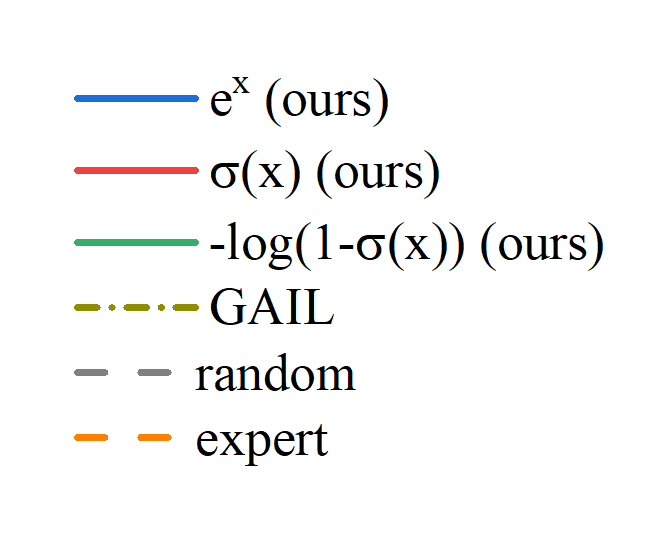}
			%			\caption{(a)}
		\end{minipage}
	}%
	
	\caption{Training curves of different algorithms for different sizes of expert trajectories. The x-axis represents the times of interaction with the environment.}
	\label{Fig_further_exp2}
\end{figure*}

\section{Experiment}

In this section, we evaluate the WDAIL approach based on the high-dimensional continuous control tasks of MuJoCo \cite{39_todorov2012mujoco}. Our Algorithm \ref{alg_1} has been implemented by PyTorch and we use the PPO algorithm with advantage estimate for improving the policy during the adversarial imitation process.

In our WDAIL method of all complex tasks, we use the neural network framework as the policy estimator which has two hidden layers of 64 units with an activation function of $tanh$, and the discriminator network has 100 units with the same activation function. We choose Adam optimizer for the policy and discriminator networks.  

To explore effective reward shapes, we conducted experiments of different reward shapes (6 kinds of reward shapes among 3 different control tasks). Besides, 3 variants of our WDAIL were compared with the popular acknowledged benchmark algorithms (BC and GAIL) and a random policy for validating the ability of imitating skills.
The results of all methods are normalized in [0,1] with respect to the performance of random and expert policy in the following comparative experiments.

\subsection{Comparison of Reward Shapes}
First, we compare different reward shapes of the WDAIL approach. The reward function is mentioned in section \ref{reward shaping}, we consider the output of the discriminator $d(s, a)$ as the input of the reward function. As you can see in the Fig.~\ref{Fig_reward_functionss}, there are two kinds of input form, which are $d(s, a)$ and $\sigma(d(s, a))$, and six different reward shapes have been constructed. The reason why we explore these kind of different reward shapes is to investigate whether the logarithmic form with sigmoid is the only reward shape effective in the adversarial imitation learning algorithm or not.
In this experiment, we evaluate the imitation performance with different 1, 5, 10, 50 trajectories and seed 0, 1, 2, 3, respectively.

The performances of WDAIL with different reward shapes and sizes of trajectories are shown in Fig.~\ref{Fig_exp1}.
\begin{itemize}
	\item $e^{x}, \sigma(x) \text { and }-\log (1-\sigma(x))$
	
	It is obvious that these three positive shapes of reward function are extremely effective for all the three tasks, such forms perform well for not only fewer trajectories but also more trajectories, which indicates that they also have great robustness.
	\item $-e^{-x}$
	
	This kind of negative function has imitated the expert data well when there is one expert trajectory, but can not perform good enough when more trajectories are provided in the experiment of HalfCheetah. In the other two environments, this shape of reward function can not imitate expert behavior at all.
	\item $x \text { and } \log (\sigma(x))$
	
	The unbiased reward function $x$ (i.e. $log(\sigma(x))-log(1-\sigma(x))$) has nice performance as the three positive reward functions behave in HalfCheetah. However, it can only achieve the poor imitation performance that random actions obtain in the other environments. Another negative reward function $log(\sigma(x))$ performs entirely the same as the unbiased function $x$ does in all these three experiments.
\end{itemize}

The training curves of this experiment are presented in Fig.~\ref{Fig_further_exp1}. These results show that our method with positive reward function may converge under 1 million interactions with the environment, which means that it has a good sample efficiency, and the learning process is stable.

All these results in this test demonstrate that we can exactly imitate the expert policy based on the positive reward of survival bonus for all three complex continuous control tasks, the unbiased reward form is suitable for two of the three tasks, but it is certain that the reward shape doesn't have to be fixed on logarithmic forms.

\subsection{Comparison with Other Algorithms}
Next, we compare our algorithm with the positive rewards to the two baseline approaches: behavior cloning (BC) and generative adversarial imitation learning (GAIL). BC and GAIL are now recognized as prevailing benchmarks of imitation learning. For GAIL, there is a little difference with WDAIL in policy network which has 100 nodes in each hidden layer, but the discriminator network is the same. There are also three high-dimensional complex environments in this experiment, and the expert demonstrations can be easily found in the baseline/gail of the OpenAI \cite{42_baselines}. 
For all the tasks, the policy of imitator has been optimized with 1, 5, 10, 50 trajectories, where each trajectory owns at most 1024 state-action pairs, and seed 0, 1, 2, 3, respectively.  

As we can see in Fig.~\ref{Fig_exp2}, WDAIL performs equally well under varying sizes of expert trajectories for almost all the environments. It is verified that BC tends to nicely imitate expert behavior when there is enough data, and GAIL presents better results than BC for most of the tasks. The learning curves in Fig.~\ref{Fig_further_exp2} displays that WDAIL acquires good performance in a small iteration for all the continuous tasks, and the learning process is smooth. 

For HalfCheetah, it is obvious that the performance has diversity for both WDAIL and GAIL providing different sizes of expert trajectories. We consider that it may be incurred by the inferior expert capability with high variance comparing to the performance of true reward in the experiment of HalfCheetah. The agent may get confused when meeting more than one inferior trajectories.

\section{Conclusion}

In this paper, we propose a new algorithm WDAIL based on the Wasserstein distance and PPO algorithm to promote the performance and sample efficiency for imitation learning. The reward shapes have been developed for making our WDAIL suitable for different high-dimensional continuous control tasks. Based on the experiment in the comparison of reward shapes, it has been verified that the logarithmic form of the reward function used in GAIL and its extensions are not the only reward shape that can make the adversarial imitation learning method work well. This property may be profit from the minimax game theory in the optimization procedure, and we believe that there is some certain interrelation and it will be explored in our future work. The results of the experiment on the comparison with other algorithms validate the superiority of our method and demonstrate that WDAIL performs great in almost all high-dimensional continuous control tasks and the training process is much more stable. 
However, WDAIL may not imitate well when there are more trajectories of inferior performance with high variance in expert demonstrations, and this will be improved in our future work.

\end{document}